\documentclass{article}
\usepackage{spconf,amsmath,graphicx,booktabs}
\usepackage[utf8]{inputenc}
\usepackage[T1]{fontenc}
\usepackage[english]{babel}
\usepackage{textcomp}
\usepackage{amsthm}
\usepackage{amsmath}
\usepackage{amssymb}
\usepackage{times}
\usepackage{soul}
\usepackage{epsfig}
\usepackage{amssymb}
\usepackage{multirow}
\usepackage{url,subfigure,graphicx,color,xcolor,booktabs,colortbl,threeparttable}
\usepackage{algorithm,algpseudocode}
\usepackage{pifont}
\usepackage{enumitem}
\usepackage{array}
\usepackage{bm}
\usepackage{url}
\usepackage{ifpdf}
\usepackage{ifxetex}
\usepackage{bbding}
\usepackage{multirow}
\usepackage{caption}
\usepackage{hyperref}
\usepackage{cite}
\newcommand{\czq}{\textcolor{black}}
\title{ProContEXT: Exploring Progressive Context Transformer for Tracking}

\vspace{-0.2in}
\name{\begin{tabular}{c}
Jin-Peng Lan$^1$\sthanks{equal contribution, alphabetically sorted},
Zhi-Qi Cheng$^2\footnotemark[1]$,
Jun-Yan He$^1$\sthanks{corresponding author}, 
Chenyang Li$^1$,
\\
\textit{Bin Luo}$^1$, 
\textit{Xu Bao}$^1$, 
\textit{Wangmeng Xiang}$^1$,
\textit{Yifeng Geng}$^1$,
\textit{Xuansong Xie}$^1$
\vspace{-0.1in}
\end{tabular}}
\address{
$^1$DAMO Academy, Alibaba Group~~~~~~~
$^2$Carnegie Mellon University ~~~~~~~\\
}

\begin{document}
%\ninept
%
\maketitle
\begin{abstract}
Existing Visual Object Tracking~(VOT) only takes the target area in the first frame as a template.
This causes tracking to inevitably fail in \textit{fast-changing} and \textit{crowded scenes}, as it cannot account for changes in object appearance between frames.
To this end, we revamped the tracking framework with \textit{\textbf{Pro}}gressive \textbf{\textit{Cont}}ext \textit{\textbf{E}}ncoding \textit{\textbf{Transformer}}
\textit{\textbf{T}}racker (\textit{\textbf{ProContEXT}}), which coherently exploits \textit{spatial and temporal contexts} to predict object motion trajectories.
Specifically, ProContEXT leverages a context-aware self-attention module to encode the spatial and temporal context, refining and updating the multi-scale static and dynamic templates to progressively perform accurately tracking.~It explores the complementary between spatial and temporal context, \textit{raising a new pathway to multi-context modeling }for transformer-based trackers.
In addition, ProContEXT revised the token pruning technique to reduce computational complexity.~Extensive experiments on popular benchmark datasets such as GOT-10k and TrackingNet demonstrate that the proposed ProContEXT achieves state-of-the-art performance\footnote{The source code is at https://github.com/jp-lan/ProContEXT}.
\end{abstract}

\vspace{-2mm}
\section{Introduction}
\vspace{-1mm}
\label{sec:intro}
\czq{Visual object tracking (VOT) is a crucial research topic due to its numerous applications, including autonomous driving, human-computer interaction, and video surveillance. In VOT, the goal is to predict the precise location of the target object in subsequent frames, given its location in the first frame, typically represented by a bounding box. However, due to challenges such as scaling and deformation, tracking systems must dynamically learn object appearance changes to encode content information. Additionally, in fast-changing and crowded scenes, visual trackers must identify which object to tracking among multiple similar instances, making tracking particularly challenging.}

\begin{figure}[t]
	\begin{center}
		\includegraphics[width=0.7\linewidth]{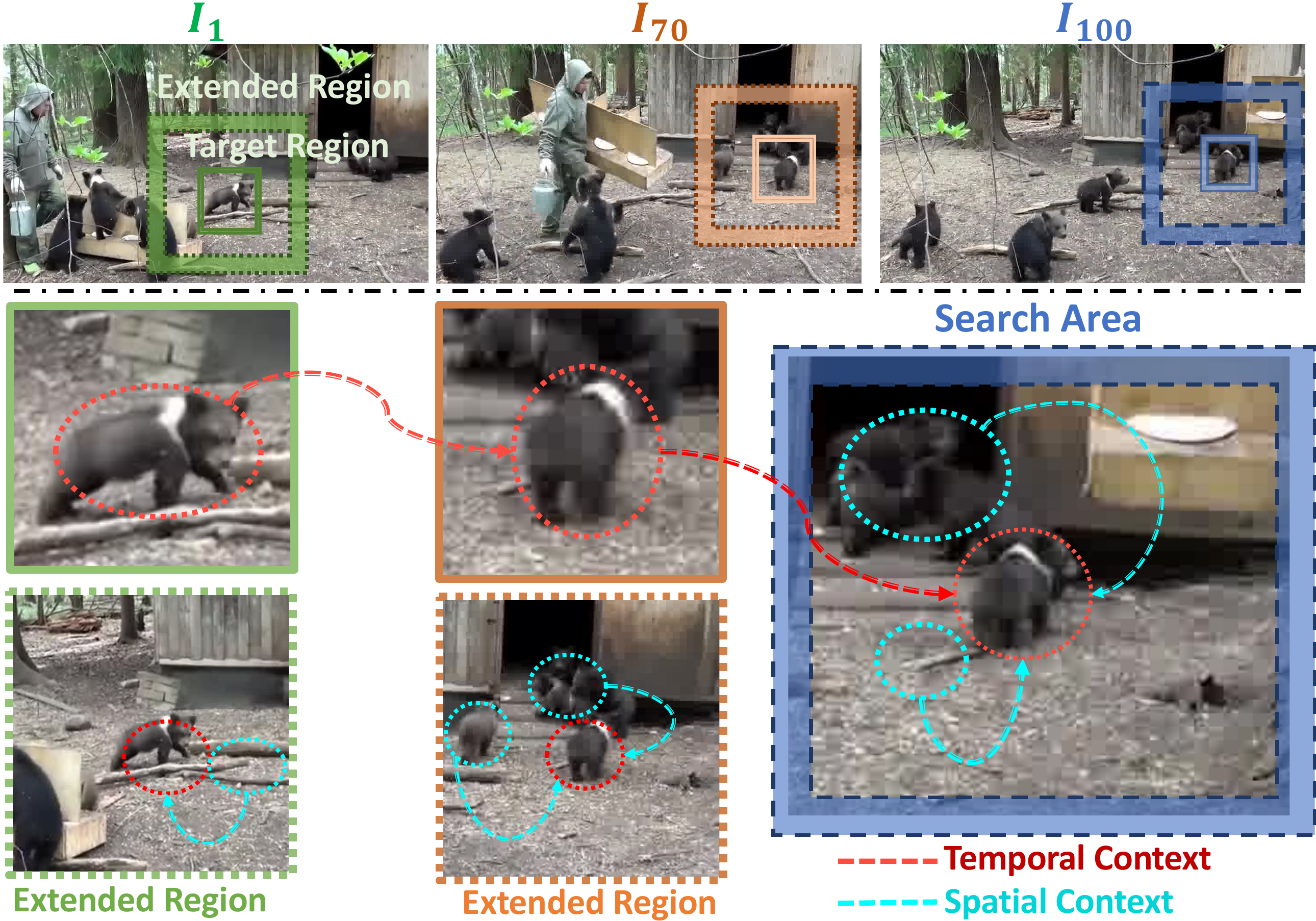}
	\end{center}
	\vspace{-0.25in}
	\caption{\footnotesize The fast-changing and crowded scenes widely exist in visual object tracking. Apparently, exploiting the \textit{temporal and spatial context} in video sequences is the cornerstone of accurate tracking.}
	\label{fig:introduction}
	\vspace{-0.25in}	
\end{figure}

\czq{To address these challenges, we propose an intuitive solution. In Fig.\ref{fig:introduction}, we show three video frames chronologically in the first row, and their cropped patches are displayed below. The middle row of Fig.\ref{fig:introduction} demonstrates that object appearance can change significantly during tracking. The red dashed circles in the middle column represent the target object, which is more similar to objects in the search area, improving tracking performance. Furthermore, at the bottom of Fig.~\ref{fig:introduction}, we extend the template regions to include more background instances marked by cyan dotted circles, which could assist trackers in identifying similar targets. Thus, temporal and spatial contexts are crucial for visual object tracking, and we refer to them as \textit{temporal context} and \textit{spatial context}, respectively.}

\begin{figure*}[t]
	\begin{center}
	\includegraphics[width=0.8\linewidth]{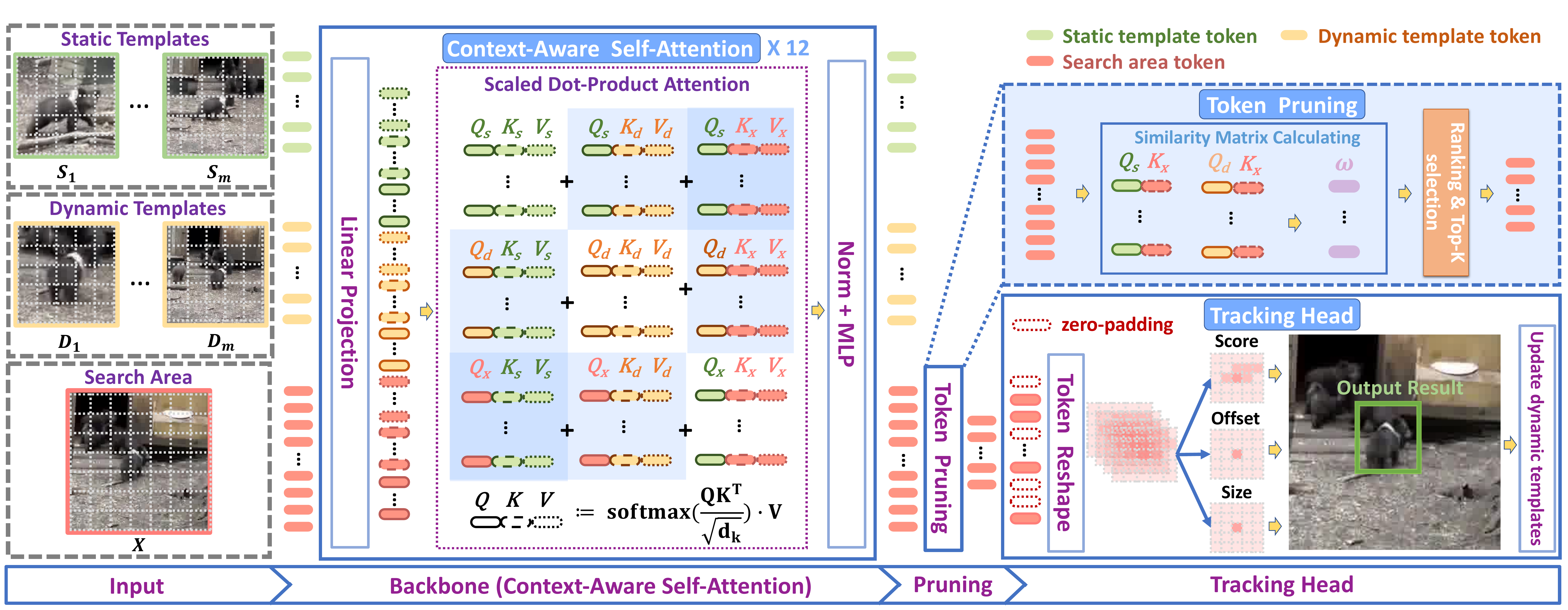}\\
	\end{center}
	\vspace{-0.25in}
	\caption{\footnotesize The framework of Progressive Context Encoding Transformer Tracker (ProContEXT).}
	\vspace{-0.25in}
	\label{fig:fig2}
\end{figure*}

Despite the emergence of context-free tracking methods, such as Siamese-based trackers (e.g., SiamFC~\cite{Bertinetto_SimaFC_ECCVW16}, SiamRPN~\cite{Li_SiamRPN_CVPR18}, and SiamRPN++\cite{Li_SiamRPN++_CVPR19}) and transformer-based approaches (e.g., TransT\cite{Chen_TrasnT_CVPR2021} and OSTrack~\cite{ye_ostrack_2022}), their performance suffers in rapidly changing scenarios due to the lack of contextual information. To address this, spatial context learning pipelines, such as TLD~\cite{Kalal_TLD_TPAMI12} and its extensions (e.g., LTCT~\cite{Ma_LTCT_CVPR15}, FuCoLoT~\cite{Lukezic_FuCoLoT_ACCV18}, and LTT~\cite{Valmadre_LTT_ECCV18}), have been developed. Furthermore, dynamic template updating has been utilized in various tasks, including perception~\cite{li2022longshortnet,cheng2022gsrformer}, segmentation~\cite{qiao2022real,he2021mgseg}, tracking~\cite{liu_MDCF_ICASSP2022, fu2021stmtrack, cui_mixformer_2022, yan_Stark_2021}, and density estimation~\cite{cheng2019learning,cheng2022rethinking}, for spatial context modeling. However, a comprehensive study of both temporal and spatial context in tracking tasks remains to be achieved.

\czq{To solve these issues, we propose a novel visual object tracking method called \textit{\textbf{Pro}}gressive \textbf{\textit{Cont}}ext \textit{\textbf{E}}ncoding \textit{\textbf{Transformer}} \textit{\textbf{T}}racker (\textit{\textbf{ProContEXT}}). ProContEXT encodes both temporal and spatial contexts through a template group composed of static and dynamic templates, providing a comprehensive and progressive context representation. The model leverages a context-aware self-attention module to learn rich and robust feature representations, while a tracking head is used to update dynamic templates and generate tracking boxes. Furthermore, we adopt token pruning techniques to improve computational efficiency without compromising performance. The main contributions of this paper are:}
\vspace{-1mm}
\czq{\begin{itemize}[leftmargin=*]
\vspace{-0.05in}
\item ProContEXT is the first work to exploit \textit{progressive context encoding} over dynamic temporal and varying spatial in transformer-based tracking.~It builds \textit{a bridge between classical contextual tracking and context-free tracking}, and investigates how to encode context in real-time tracking.
\vspace{-0.05in}
\item ProContEXT revises the ViT backbone, adding more \textit{static and dynamic templates} and improving \textit{context-aware self-attention} to exploit multi-temporal and multi-spatial information.~With \textit{progressive template refinement and updates}, it alters the \textit{token pruning} to seamlessly bring contextual encoding into transformer-based tracking.
\vspace{-0.05in}
\item ProContEXT \textit{achieves SOTA performance} on large scale tracking benchmarks including TrackingNet~\cite{muller_trackingnet_ECCV2018} and GOT-10k~\cite{Huang_GOT10K_TPAMI21}.~Despite expanding long-term temporal and multi-scale spatial information, ProContEXT can \textit{perform context encoding and tracking in real-time} at 54.3 FPS.
\end{itemize}}
\vspace{-0.2in}

\vspace{-4mm}
\czq{\section{Methodology}
\label{sec:method}
\vspace{-0.1in}
\subsection{Network Architecture}
\vspace{-0.1in}
Unlike most previous works~\cite{Li_SiamRPN_CVPR18,Chen_TrasnT_CVPR2021,ye_ostrack_2022} that use only static templates at the first frame, ProContEXT aims to exploit templates of \textit{multi-temporal and multi-spatial} to encode more context information.}

\noindent \textbf{Static \& Dynamic Templates}.~As shown in Fig.~\ref{fig:fig2}, given video frames {\small $\mathcal{V}$=$[I_1,...,I_n]$} and an initial box {\small $b_{init}$}, we crop the templates at different scales {\small $\mathcal{K}$}={\small $\{k_1, \dots, k_m\}$} to generate static templates {\small $\mathcal{S}$=$\{S_{1}, \dots, S_{m}\}$}, where $S_t$ is cropped at scale {\small $k_t$}.~Similarly, the dynamic templates {\small $\mathcal{D}$=$\{D_{1},\dots, D_{m}\}$} is produced to encode more object appearance changes during tracking.~Following the standard setting~\cite{ye_ostrack_2022,cui_mixformer_2022,yan_Stark_2021,cheng2017video2shop}, we also crop the area centered on the box at the previous frame to get search area {\small $\mathcal{X}$}, assuming that the target object appears in the adjacent region of the last known location.\footnote{We're abusing notation and symbolizing search areas and templates in the same way for ease of introduction.}~Overall, we expand the dynamic templates with multi-temporal and multi-spatial details, which is plainly different from previous works that only use one static template.

\noindent \czq{\textbf{Context-Aware Self-Attention}. Based on the expanded dynamic templates, we modify the ViT~\cite{dosovitskiy_ViT_2020} for representation learning. First, all templates {\small $\mathcal{S} \cup \mathcal{D}$} and the search area {\small $\mathcal{X}$} are fed into a rescaling module for resizing. Then, each resized patch is cropped into non-overlapping {\small $16 \times 16$} image patches, flattened to 1D, and positional embeddings are added after passing through the patch embedding layer. In the end, we encode {\small $\mathcal{S}$}, {\small $\mathcal{D}$}, and {\small $\mathcal{X}$} into static tokens {\small $\mathcal{Z}_s$=$\{Z_s^{1};\dots;Z_s^{m}\}$}, dynamic tokens {\small $\mathcal{Z}_d$=$\{Z_d^{1};\dots;Z_d^{m}\}$}, and search tokens {\small $\mathcal{Z}_x$=${Z_x}$}, respectively. Then, all tokens are concatenated as {\small $[\mathcal{Z}_s;\mathcal{Z}_d;\mathcal{Z}_x]$} and input coherently into the self-attention module as,}
\begin{equation}
\vspace{-2mm}
\scriptsize
	\label{eq1}
	\begin{split}
		\text{A}(Q,K,V) &= \text{Softmax}(\frac{QK^T}{\sqrt{d_k}})\cdot V
		\\&=\text{Softmax}(\frac{[Q_s;Q_d;Q_x][K_s;K_d;K_x]^T}{\sqrt{d_k}})\cdot [V_s;V_d;V_x]
		\\&=\text{Softmax}(\frac{\begin{bmatrix}
				Q_sK_s^T & Q_sK_d^T & Q_sK_x^T \\
				Q_dK_s^T & Q_dK_d^T & Q_dK_x^T \\
				Q_xK_s^T & Q_xK_d^T & Q_xK_x^T \\
		\end{bmatrix}}{\sqrt{d_k}})\cdot \begin{bmatrix}V_s\\V_d\\V_x	\end{bmatrix}
	\end{split}
\end{equation}
\czq{where Eq.~(\ref{eq1}) shows that the context-aware self-attention module can encode both dynamic temporal and variable spatial information from all static and dynamic templates. Diagonal terms, such as {\small $Q_xK_x^T$}, focus on the intra-region representation of the search area, while {\small $Q_sK_s^T$} and {\small $Q_dK_d^T$} fuse the spatial context in static templates and dynamic templates, respectively. Off-diagonal terms, such as {\small $Q_sK_x^T$} and {\small $Q_dK_x^T$}, account for the interactions between templates and the search area. {\small $Q_xK_s^T$} and {\small $Q_xK_d^T$} aggregate the temporal context into the search areas, while {\small $Q_sK_d^T$} and {\small $Q_dK_s^T$} denote inter-template interactions. By leveraging 12 stacked attention layers, ProContEXT progressively extracts context-aware features.}

\noindent \textbf{Tracking Head}.~The tracking head consists of score head, offset head, and size head.~After representation learning, only the tokens of the search area {\small $\mathcal{X}$} are reshaped into the 2D feature ({\small $W_x \times H_x$}), and fed into the tracking head.~The reshaping here aims to convert 1D flattened tokens into the spatial domain to perform the same tracking process as other convolutional networks. Specifically, the score head first gets a rough position and a score map. Then offset head and size head refine the offsets of position and box size on the resulting score map, respectively. More details are in Eq.~(\ref{eq:g-x-y}-\ref{eq:final-loss} and corresponding descriptions.

\noindent \czq{\textbf{Token Pruning}.~We modified the token pruning technique from previous works~\cite{ye_ostrack_2022,rao_dynamicvit_NeurIPS2021} to accelerate ProContEXT. The purpose of pruning is to reduce computational costs by ignoring the search tokens for noisy background patches. Unlike OSTrack~\cite{ye_ostrack_2022}, which only considers the static template of the first frame, we take into account all static and dynamic templates. This allows foreground tokens to remain similar to dynamic templates even after significant changes in appearance. Moreover, the target object is usually located at the center point of the templates. If the similarity between the center point and the search token is low, the search token can be determined as the background. According to Eq.~(\ref{eq1}), the score of the search area token is computed by {\small $\omega=\phi(\text{softmax}(Q_sK_x^T/\sqrt{d_k})+\text{softmax}(Q_dK_x^T/\sqrt{d_k})) \in\mathbb{R}^{1\times N_x}$}, where {\small $N_x$} is the number of search tokens, and {\small $\phi()$} sums up the attention matrix that bonds to the template center tokens.~Finally, we only keep the search tokens for the top-k elements of {\small $\omega$}.~The pruned tokens are replaced with zero-padding and then fed to the tracking head.}

\vspace{-0.15in}
\subsection{Training and Inference Settings}
\vspace{-2mm}
\czq{After presenting the architecture of ProContEXT, we now describe the training and inference settings.}

\noindent \textbf{Training Settings}.~\czq{Inspired by previous works~\cite{zhou2019objects}, the training process is a progressive optimization process. First, the score head predicts the approximate location and score of the target object, where the Gaussian kernel generates the supervision as,}
\begin{equation}
	\footnotesize
	\centering
	G_{xy}=\exp(-\frac{(x-p_x)^2+(y-p_y)^2}{2\sigma_p^2}).
	\label{eq:g-x-y}
\end{equation}
where {\small $(p_x, p_y)$} are the coordinates of the center point, and {\small $\sigma_p$} is the standard deviation that defines the size of the object.
With the supervision of the Gaussian kernel, the score head is optimized with the focal loss~\cite{lin_focal_iccv2017},
\begin{equation}
	\footnotesize
	\centering
	L_{s} = -\Sigma\begin{cases}
		(1-\hat{G}_{xy})^\alpha \log(\hat{G}_{xy}),  \;\;  \text{if}  \; G_{xy} = 1 \\
		(1-G_{xy})^\beta (\hat{G}_{xy})^\alpha \log(1-\hat{G}_{xy}), \;\; \text{otherwise}
	\end{cases}
	\label{eq:focal-loss}
\end{equation}
where {\small $\hat{G}_{xy}\in[0,1]^{W_x\times H_x}$} is the score map, ({\small $W_x$, $H_x$}) is the feature size of search area. We set {\small $\alpha=2$} and {\small $\beta=4$} as previous work~\cite{zhou2019objects}. After obtaining the maximum response of the score head {\small $(\hat{x}, \hat{y}) = \text{argmax}_{x,y}(\hat{G}_{xy})$}, the final predicted box can be computed as,
\begin{equation}
	\footnotesize
	\centering
	\hat{b} = (\hat{x}+\hat{\delta}_{x}, \hat{y}+\hat{\delta}_{y}, \hat{w}, \hat{h})
	\label{eq:final-box}
\end{equation}
where {\small $(\hat{\delta}_{x}, \hat{\delta}_{y})$} is the offset from the offset head, and {\small $(\hat{w},\hat{h})$} is the box size from the size head at location {\small $(\hat{x}, \hat{y})$}, respectively. 
The bounding box obtained in Eq.~\ref{eq:final-box} is trained with IoU loss~\cite{rezatofighi_IoULoss_CVPR2019} and L1 loss. Finally, the total loss can be noted as,
\begin{equation}
    \footnotesize
    \centering
	Loss= L_{s} + \lambda_{iou}L_{iou}(\hat{b}, b_{gt}) + \lambda_{l_1}L_1(\hat{b}, b_{gt})
	\label{eq:final-loss}
\end{equation}
where $\lambda_{iou}=2$ and $\lambda_{l_1}=5$ are the loss weights as in~\cite{yan_Stark_2021}. More details are in Sec.~\ref{sec:implementation}.

\noindent \textbf{Inference Details}.~Unlike MixFormer~\cite{cui_mixformer_2022} and STARK~\cite{yan_Stark_2021} that use an extra branch to update templates, we reuse the score head in inference. We take the maximum response of the score head as a confidence score,
\begin{equation}
	\footnotesize
	\centering
	\text{score} = \max(\hat{G}_{xy})
	\label{eq:final-score}
\end{equation}
where Alg.~\ref{alg:cap} depicts how to update the template in inference.
After initializing static and dynamic templates (Line~5), confidence scale and position are used to assume whether to update the dynamic template (Line~8-9).
Supposing that the confidence score is above a threshold {\small $\tau$} (Line~10), the current tracking result {\small $b_{pred}$} is regarded as reliable and utilized to update the multi-scale dynamic templates progressively (Line~11).

\begin{algorithm}[!t]
\footnotesize
\caption{\footnotesize Inference with updating templates}
\label{alg:cap}
\begin{algorithmic}[1]
\State \textbf{Input:} video frames $\mathcal{V}=[I_1,...,I_n]$, initial box $b_{init}$, scales $\mathcal{K}=[k_1,...,k_m]$, and trained $\text{ProContEXT}()$;
\State \textbf{Output:} box for each frame $\mathcal{B}=[b_1,...,b_n]$;
\State \textbf{For} each frame $I_{i}$ in $\mathcal{V}$ \textbf{do}
    \State \hspace{\algorithmicindent} \textbf{if} $i == 1$ \textbf{then}

       \State \hspace{\algorithmicindent} \hspace{\algorithmicindent} Initialize templates $\mathcal{S} = \text{crop}(I_1, b_{init}, \mathcal{K})$, $\mathcal{D} = \mathcal{S}$;
		\State \hspace{\algorithmicindent} \hspace{\algorithmicindent} Initialize predicted box $b_{pred} =  b_{init}$; 
		\State \hspace{\algorithmicindent} \textbf{else}
            \State \hspace{\algorithmicindent} \hspace{\algorithmicindent}  
            Determine a search area $\mathcal{X} = \text{crop}(I_i, b_{pred})$;
  \State \hspace{\algorithmicindent} \hspace{\algorithmicindent} $b_{pred}, score = \text{ProContEXT}(\mathcal{S}, \mathcal{D}, \mathcal{X})$;  \Comment{Eq. (\ref{eq:final-box},\ref{eq:final-score})}
		\State \hspace{\algorithmicindent} \hspace{\algorithmicindent} \textbf{if} $score > \tau$  \textbf{then} %\Comment{update dynamic templates}
		\State \hspace{\algorithmicindent} \hspace{\algorithmicindent}\hspace{\algorithmicindent} $\mathcal{D}=\{D_{1},D_{2}, ...,D_{m}\}=\text{crop}(I_i,b_{pred},\mathcal{K})$; 
		\State \hspace{\algorithmicindent} \hspace{\algorithmicindent} \textbf{end if}
		\State \hspace{\algorithmicindent} \textbf{end if}
            \State \hspace{\algorithmicindent}
            Update outputs   $\mathcal{B}[i]=b_{pred}$;
		\State\textbf{end for}
		\State \textbf{return} $\mathcal{B}=[b_{1},\dots,b_{n}]$;
	\end{algorithmic}

\end{algorithm}

\vspace{-0.1in}
\section{Experiments}

\label{sec:experiment}

\subsection{Dataset and Metric}
\vspace{-0.1in}

\czq{To conduct a thorough evaluation of the proposed ProContEXT, we utilize large-scale VOT benchmark datasets: TrackingNet~\cite{muller_trackingnet_ECCV2018} and GOT-10k~\cite{Huang_GOT10K_TPAMI21}. The GOT-10k dataset includes more than 10,000 video segments and over 1.5 million manually labeled bounding boxes, while TrackingNet consists of 30,000 sequences with 14 million annotations. We adhere to the evaluation protocols of these datasets, with the average overlap (AO) and success rate (SR) used for evaluation on the train split of GOT-10k, and the area under the curve (AUC), precision, and normalized precision used for evaluation on TrackingNet. We also employ the same training datasets as recent state-of-the-art methods~\cite{yan_Stark_2021,cui_mixformer_2022,ye_ostrack_2022} when evaluating on TrackingNet.}

\vspace{-0.1in}
\subsection{Implementation Details}
\vspace{-0.1in}
\label{sec:implementation}
\czq{We implement the proposed ProContEXT using PyTorch and utilize the ViT-base backbone that is pre-trained by MAE~\cite{MaskedAutoencoders2021}. Token pruning is conducted before the 4th, 7th, and 10th blocks at a keeping ratio of 0.7, which is consistent with~\cite{ye_ostrack_2022,rao_dynamicvit_NeurIPS2021}. For training, we set the batch size and learning rate to 128 and 1e-4, respectively, and train the model using the AdamW solver for 300 epochs. We employ horizontal flip and template jittering in scale and size for data augmentations. All templates are resized to {\small $192\times192$}, and the search area is resized to {\small $384\times384$} for ProContEXT. During inference, the score threshold {\small $\tau$} is set to 0.7. All experiments are conducted using 4 NVIDIA A100 GPUs.}

\begin{table} %[ht]
\centering
\footnotesize
\caption{\footnotesize 
\czq{Comparison of State-of-the-Art (SOTA) Methods on TrackingNet~\cite{muller_trackingnet_ECCV2018} and GOT-10k~\cite{Huang_GOT10K_TPAMI21} Datasets: The evaluated methods include "DT" (Dynamic Template) and "EB" (Extra Branch to update Dynamic Templates), along with different initialization methods, such as "Random-None" and pre-training with additional datasets, such as "CLIP-WIT\cite{radford_CLIP_ICML2021}", "CLS-ImageNet-1k\cite{deng_imagenet_CVPR2009}", "CLS-ImageNet-22k\cite{ridnik_imagenet21k_2021}", or "MAE-ImageNet-1k\cite{MaskedAutoencoders2021}". The best performance is denoted in \textcolor[RGB]{0,153,51}{green} font, and the corresponding performance increase is displayed in \textcolor[RGB]{255,0,0}{red} font. When reporting results on the GOT-10k, trackers were only trained on the GOT-10k training split.
}}
\vspace{-2mm}
	\resizebox{\linewidth}{!}{
		\begin{tabular}{c |c | c| c | c c c | c c c}
			% \hlinew{1pt} \hlinew{1pt}
			\hline
			\multirow{2}*{\textbf{Method}} & 
			% \multirow{2}*{\textbf{PTMtd}} & 
                \textbf{Pre-training} & 
			% \multirow{2}*{\textbf{PTData}} &
			\multirow{2}*{\textbf{DT}} & 
			\multirow{2}*{\textbf{EB}} 	& 
			\multicolumn{3}{c}{\textbf{TrackingNet}}\vline 			 &  \multicolumn{3}{c}{\textbf{GOT-10k}} \\
			% ~	&~	&~ & ~		& ~	&  \textbf{AUC} & \textbf{N-PRE} & 
                ~	& \textbf{Method-Data} &~		& ~	&  \textbf{AUC} & \textbf{N-PRE} &
   \textbf{PRE} & \textbf{AO} &\boldsymbol{$SR_{0.5}$} &  \boldsymbol{$SR_{0.75}$} \\ \hline
SiamFC~\cite{Bertinetto_SimaFC_ECCVW16} & Random-None &$\times$ & -   & 57.1 & 66.3 & 53.3 					& 34.8 & 35.3 & 9.8 \\
SiamRPN++~\cite{Li_SiamRPN++_CVPR19} & Random-None &$\times$ & -  & 73.3 & 80.0 & 69.4 & 51.7 & 61.6 & 32.5 \\
SimTrack~\cite{chen_SimTrack_2022}   &CLIP-WIT &$\times$& -	& 82.3	& 86.5 & -  					& 68.6 & 78.9 & 62.4 \\
\hline
Stmtrack~\cite{fu2021stmtrack}    &CLS-ImageNet-1k     & $\times$& -    & 80.3 & 85.1 & 76.7 					& 64.2 & 73.7 & 57.5 \\
TransT~\cite{Chen_TrasnT_CVPR2021} &CLS-ImageNet-1k	& $\times$& -	& 81.4 & 86.7 & 80.3 					& 67.1 & 76.8 & 60.9 \\
STARK~\cite{yan_Stark_2021}  &CLS-ImageNet-1k  	   	&  \checkmark & \checkmark 	  & 82.0 & 86.9 & - &68.8  & 78.1 & 64.1 \\
MixFormer~\cite{cui_mixformer_2022} &CLS-ImageNet-1k &  \checkmark &  \checkmark 	& 82.6 & 87.7 & 81.2 & 71.2 & 79.9 & 65.8 \\
\multirow{2}*{ProContEXT}&\multirow{2}*{CLS-ImageNet-1k}& \multirow{2}*{\textbf{\checkmark}} &  \multirow{2}*{\textbf{$\times$}} & \textbf{\textcolor[RGB]{0,153,51}{83.4}} & \textbf{\textcolor[RGB]{0,153,51}{88.3}} & \textbf{\textcolor[RGB]{0,153,51}{82.2}} & \textbf{\textcolor[RGB]{0,153,51}{72.0}} & \textbf{\textcolor[RGB]{0,153,51}{82.7}} & \textbf{\textcolor[RGB]{0,153,51}{68.0}}  \\
~                &~ 	&~	  & ~   	& $\uparrow$ \textbf{\textcolor[RGB]{255,0,0}{+0.8}} & $\uparrow$ \textbf{\textcolor[RGB]{255,0,0}{+0.6}} &  $\uparrow$ \textbf{\textcolor[RGB]{255,0,0}{+1.0}} & $\uparrow$ \textbf{\textcolor[RGB]{255,0,0}{+0.8}} &  $\uparrow$ \textbf{\textcolor[RGB]{255,0,0}{+2.8}} &  $\uparrow$ \textbf{\textcolor[RGB]{255,0,0}{+2.2}} \\
\hline
MixFormer~\cite{cui_mixformer_2022} &CLS-ImageNet-22k &  \checkmark &  \checkmark 	& 83.1 & 88.1 & 81.6 					& 70.7 & 80.0 & 67.8 \\
\multirow{2}*{ProContEXT}&\multirow{2}*{CLS-ImageNet-22k}& \multirow{2}*{\textbf{\checkmark}} &  \multirow{2}*{\textbf{$\times$}} & \textbf{\textcolor[RGB]{0,153,51}{83.2}} & \textbf{\textcolor[RGB]{0,153,51}{88.0}} & \textbf{\textcolor[RGB]{0,153,51}{81.7}} & \textbf{\textcolor[RGB]{0,153,51}{72.5}} & \textbf{\textcolor[RGB]{0,153,51}{82.6}} & \textbf{\textcolor[RGB]{0,153,51}{69.3}} \\
~                &~ 	&~	  & ~   	& $\uparrow$ \textbf{\textcolor[RGB]{255,0,0}{+0.1}} & $\downarrow$ \textbf{\textcolor[RGB]{255,0,0}{-0.1}} &  $\uparrow$ \textbf{\textcolor[RGB]{255,0,0}{+0.1}} & $\uparrow$ \textbf{\textcolor[RGB]{255,0,0}{+1.8}} &  $\uparrow$ \textbf{\textcolor[RGB]{255,0,0}{+2.6}} &  $\uparrow$ \textbf{\textcolor[RGB]{255,0,0}{+1.5}} \\
% ProContEXT & CLS-ImageNet-22k &  \checkmark &  $\times$ & ~ & ~ & ~ & 72.5 & 82.6 & 69.3 \\
\hline
OSTrack~\cite{ye_ostrack_2022}	 &MAE-ImageNet-1k 	& $\times$& -	& 83.9 & 88.5 & 83.2 						 & 73.7 & 83.2 & 70.8 \\
\multirow{2}*{ProContEXT}   &\multirow{2}*{MAE-ImageNet-1k}	 & \multirow{2}*{\textbf{\checkmark}} & \multirow{2}*{\textbf{$\times$}}   	& \textbf{\textcolor[RGB]{0,153,51}{84.6}} & \textbf{\textcolor[RGB]{0,153,51}{89.2}} &  \textbf{\textcolor[RGB]{0,153,51}{83.8}} & \textbf{\textcolor[RGB]{0,153,51}{74.6}} & \textbf{\textcolor[RGB]{0,153,51}{84.7}} & \textbf{\textcolor[RGB]{0,153,51}{72.9}}   \\
~              &~    & ~ & ~   	& $\uparrow$ \textbf{\textcolor[RGB]{255,0,0}{+0.7}} & $\uparrow$ \textbf{\textcolor[RGB]{255,0,0}{+0.7}} &  $\uparrow$ \textbf{\textcolor[RGB]{255,0,0}{+0.6}} & $\uparrow$ \textbf{\textcolor[RGB]{255,0,0}{+0.9}} &  $\uparrow$ \textbf{\textcolor[RGB]{255,0,0}{+1.5}} &  $\uparrow$ \textbf{\textcolor[RGB]{255,0,0}{+2.1}}    \\\hline
	\end{tabular}}
	\label{tab:comparing-sota}
	\vspace{-0.15in}
\end{table}

\vspace{-0.15in}
\subsection{Comparison with State-of-the-Art Methods}
\vspace{-0.1in}
\begin{table} \scriptsize 
    \centering
	\caption{\footnotesize 
 \czq{Number of Scales for Static Templates: The best result is denoted in \textcolor[RGB]{0,153,51}{green} font, and the corresponding increase is displayed in \textcolor[RGB]{255,0,0}{red} font.}}
	\vspace{-0.1in}
	\setlength{\tabcolsep}{1.35mm}{
		\begin{tabular}{c|c | c |c c}
			% \hlinew{1pt} \hlinew{1pt}
			\hline
			\multirow{2}*{\textbf{Method}} & \textbf{scale}  & \multirow{2}*{\boldsymbol{$\mathcal{K}$}} & \multicolumn{2}{c}{\textbf{GOT-10k(val)}} \\ %\cline{4-5}
			~ 													&  \textbf{num} & ~ & \textbf{AO} & \boldsymbol{$SR_{0.5}$} \\ \hline
			\multirow{4}*{ProContEXT}                           & 1        &    [2.0]    & 85.1 & 94.6            \\
			~                           &2         & [2.0, 4.0]   & \textbf{\textcolor[RGB]{0,153,51}{86.7}} $\uparrow$  \textbf{\textcolor[RGB]{255,0,0}{+1.6}} &                \textbf{\textcolor[RGB]{0,153,51}{96.5}} $\uparrow$ \textbf{\textcolor[RGB]{255,0,0}{+1.9}}            \\
			~                          & 3          &  [2.0, 3.0, 4.0]  & 85.4 $\uparrow$ +0.3 & 95.6 $\uparrow$ +1.0         \\ 
			~                          & 4         & [2.0, 2.7, 3.3, 4.0] & 85.4 $\uparrow$ +0.3 & 95.4  $\uparrow$ + 0.8       \\ \hline
	\end{tabular}}
	\vspace{-0.2in}
	\label{table:scales}
\end{table}

\czq{The performance comparison of the proposed ProContEXT with state-of-the-art methods on TrackingNet~\cite{muller_trackingnet_ECCV2018} and GOT-10k~\cite{Huang_GOT10K_TPAMI21} is presented in Table~\ref{tab:comparing-sota}.
When pre-training with the ImageNet-1k dataset, ProContEXT achieves the best results among all compared methods. By utilizing MAE pre-training, ProContEXT outperforms the recent SOTA method, OStrack~\cite{ye_ostrack_2022}, by 0.9\%, 1.5\%, and 2.1\% for AO, SR${0.5}$, and SR${0.75}$ on GOT-10k~\cite{Huang_GOT10K_TPAMI21}, respectively.
Furthermore, ProContEXT shows superior performance compared to methods~\cite{yan_Stark_2021,cui_mixformer_2022} that utilize an extra branch to update dynamic templates, while ProContEXT requires no additional branch and computation for updating dynamic templates.
To validate the generalization ability of ProContEXT, we also evaluate our model on the TrackingNet~\cite{muller_trackingnet_ECCV2018} dataset, and it demonstrates better performance than the SOTA methods.}

\vspace{-0.1in}
\subsection{Ablation Study}
\vspace{-0.1in}
\czq{We conduct an ablation study on the GOT-10k~\cite{Huang_GOT10K_TPAMI21} benchmark. We rescale all the templates and the search area to {\small $128\times 128$} and {\small $256\times 256$}, respectively. We train ProContEXT for 100 epochs on the training split and report its performance on the validation split.}
\label{exp:scale}

\begin{table}[!ht]
\scriptsize 
\centering
\caption{\footnotesize 
\czq{Exploration of Dynamic Templates: The best result is denoted in \textcolor[RGB]{0,153,51}{green} font, and the corresponding increase is displayed in \textcolor[RGB]{255,0,0}{red} font.}
}
\vspace{-0.1in}
\setlength{\tabcolsep}{1.35mm}{
\begin{tabular}{c|c | c | c c}
% \hlinew{1pt} \hlinew{1pt}
\hline
\multirow{2}*{\textbf{Method}}      & \multirow{2}*{\boldsymbol{$\mathcal{K}$}} & \textbf{Dynamic} & \multicolumn{2}{c}{\textbf{GOT-10k(val)}} \\ %\cline{4-5}
~ 							        & ~ & \textbf{template} & \textbf{AO} & \boldsymbol{$SR_{0.5}$} \\ \hline
\multirow{4}*{ProContEXT}                           &     [2.0]     &           & 85.1 & 94.6                        \\ %\hline
~                                   &      [2.0]    & \Checkmark         & 86.3 $\uparrow$ +1.2 & 96.0 $\uparrow$ +1.4            \\ %\cline{2-5}
% 			~                           & \Checkmark       &          & 86.7 $\uparrow$ +1.6 & 96.5 $\uparrow$ +1.9             \\ %\hline

~                                   & [2.0, 4.0]          & \Checkmark         & \textbf{\textcolor[RGB]{0,153,51}{86.8}} $\uparrow$ \textbf{\textcolor[RGB]{255,0,0}{+1.7}}  &  \textbf{\textcolor[RGB]{0,153,51}{96.8}} $\uparrow$ \textbf{\textcolor[RGB]{255,0,0}{+2.2}}         \\ \hline
\end{tabular}}
\vspace{-0.1in}
\label{table:ablation}
\end{table}

\noindent \czq{\textbf{Investigation of Templates:} To validate the effectiveness of spatial and temporal context, we analyzed the performance of static and dynamic templates. For static templates, we investigated scale factors $\mathcal{K}$ and scale numbers. The minimum scale factor of the template was set to 2, as in previous works~\cite{ye_ostrack_2022,yan_Stark_2021}. The maximum scale factor of the template was set to 4, which is the same as the search area. For different scale factor numbers, we uniformly distributed the scale factors between 2 and 4. From the results presented in Table~\ref{table:scales}, we observed that all models with multi-scale static templates enhanced performance since ProContEXT can effectively encode spatial context. Notably, ProContEXT with only one additional static template achieved the best result (+1.6\% in AO and +1.9\% in SR), suggesting that more static templates may introduce additional noise.}

\czq{We also performed experiments to investigate the contribution of dynamic templates. From Table~\ref{table:ablation}, it was evident that adding dynamic templates improved performance by 1.2\% in AO and 1.4\% in SR, demonstrating the effectiveness of temporal context. Furthermore, ProContEXT with both multi-scale static and dynamic templates achieved more than 2\% improvement in SR, demonstrating the complementary nature of spatial and temporal context.}

\begin{table}
\scriptsize 
\centering
\caption{\footnotesize 
\czq{Exploration of Keeping Ratio $\rho$ in Token Pruning: The best results and improvements are denoted in \textcolor[RGB]{0,153,51}{green} and \textcolor[RGB]{255,0,0}{red} font, respectively.}
}
\vspace{-0.1in}
\setlength{\tabcolsep}{1.35mm}{
\begin{tabular}{c|c | c | c c}
% \hlinew{1pt} \hlinew{1pt}
\hline
\multirow{2}*{\textbf{Method}}      &\multirow{2}*{\boldsymbol{$\rho$}} & \multirow{2}*{\textbf{GFLOPs}} & \multicolumn{2}{c}{\textbf{GOT-10k(val)}}\\ %\cline{4-5}
~ 									& ~		    & ~ & \textbf{AO} & \boldsymbol{$SR_{0.5}$}  \\ \hline
\multirow{5}*{ProContEXT}  & 0.6       & \textbf{\textcolor[RGB]{0,153,51}{36.1}} $\downarrow$ \textbf{\textcolor[RGB]{255,0,0}{20.7\%}}  & 85.8 $\uparrow$ +0.1 & 95.4    $\uparrow$ -0.1                    \\
~                         			& 0.7       & 38.0 $\downarrow$ 16.5\%  & \textbf{\textcolor[RGB]{0,153,51}{86.8}} $\uparrow$ \textbf{\textcolor[RGB]{255,0,0}{+1.1}} & \textbf{\textcolor[RGB]{0,153,51}{96.8}}    $\uparrow$ \textbf{\textcolor[RGB]{255,0,0}{+1.3}}    \\
~                          			& 0.8       & 40.1 $\downarrow$ 11.9\%  & 86.3 $\uparrow$ +0.3 & 96.1   $\uparrow$ +0.6       \\
~                          			& 0.9       & 42.7 $\downarrow$ 6.2\%   & 86.3 $\uparrow$ +0.6 & 95.8  $\uparrow$ +0.3          \\
~                          			& 1.0       & 45.5                      & 85.7                 & 95.5   \\ \hline
\end{tabular}}
\label{table:pruning}
\vspace{-0.2in}
 \end{table}

\noindent \czq{\textbf{Effects of Token Pruning:} Table~\ref{table:pruning} presents the effects of the token pruning module. The keeping ratio {\small $\rho$} indicates the proportion of reserved tokens after the pruning operation. As the value of {\small $\rho$} decreases, the computation amount decreases as well. When {\small $\rho$} is set to 0.7, ProContEXT improves AO by 1.1\% and reduces GFLOPs by 16.5\%, achieving a good trade-off between computation cost and accuracy.}

\vspace{-0.1in}
\section{Conclusion}
\label{sec:pagestyle}

\czq{In this work, we proposed the Progressive Context Encoding Transformer Tracker (ProContEXT) to revamp the visual object tracking framework. ProContEXT utilizes a context-aware self-attention module to encode spatial and temporal context, progressively refining and updating multi-scale static and dynamic templates for accurate tracking. Our extensive experiments on popular benchmark datasets such as GOT-10k and TrackingNet demonstrate that ProContEXT achieves state-of-the-art performance. In the future, we plan to explore more effective context-learning strategies and token-pruning schemes to reduce the impact of complex contexts.}

\vfill\pagebreak
{\small \bibliographystyle{IEEEbib}
\bibliography{refs_sim}}

\end{document}